\documentclass[conference]{IEEEtran}
\IEEEoverridecommandlockouts
\IEEEpubid{\makebox[\columnwidth]{DOI: 10.1109/ICDM54844.2022.00167 ©2022 IEEE \hfill} \hspace{\columnsep}\makebox[\columnwidth]{ }}
\usepackage{cite}
\usepackage{amsmath,amssymb,amsfonts}
\usepackage[ruled,vlined]{algorithm2e}

\usepackage{array}
\usepackage{multirow}
\usepackage{url}

\usepackage{graphicx}
\usepackage[export]{adjustbox}

\usepackage{textcomp}
\usepackage[dvipsnames]{xcolor}
\def\BibTeX{{\rm B\kern-.05em{\sc i\kern-.025em b}\kern-.08em
    T\kern-.1667em\lower.7ex\hbox{E}\kern-.125emX}}
    
\usepackage{microtype}
\usepackage{subfigure}
\usepackage{mathtools}
\usepackage{arydshln}
\usepackage{amsthm}

\newtheorem{definition}{Definition}

\newcommand\scalemath[2]{\scalebox{#1}{\mbox{\ensuremath{\displaystyle #2}}}}

\begin{document}

\title{Uncertainty Propagation in Node Classification
\thanks{The work was supported by H2020 MonB5G project (grant agreement no. 871780).}
}

\author{\IEEEauthorblockN{Zhao Xu\textsuperscript{\dag}, Carolin Lawrence\textsuperscript{\dag}, Ammar Shaker\textsuperscript{\dag}, Raman Siarheyeu\textsuperscript{\dag}}
\IEEEauthorblockA{\textsuperscript{\dag}{NEC Laboratories Europe, Heidelberg, Germany} \\
\{Zhao.Xu, Carolin.Lawrence, Ammar.Shaker, Raman.Siarheyeu\}@neclab.eu}
}

\maketitle

\begin{abstract}
Quantifying predictive uncertainty of neural networks has recently attracted increasing attention. In this work, we focus on measuring uncertainty of graph neural networks (GNNs) for the task of node classification. Most existing GNNs model message passing among nodes. The messages are often deterministic. Questions naturally arise: Does there exist uncertainty in the messages? How could we propagate such uncertainty over a graph together with messages? To address these issues, we propose a Bayesian uncertainty propagation (BUP) method, which embeds GNNs in a Bayesian modeling framework, and models predictive uncertainty of node classification with Bayesian confidence of predictive probability and uncertainty of messages. Our method proposes a novel uncertainty propagation mechanism inspired by Gaussian models. Moreover, we present an uncertainty oriented loss for node classification that allows the GNNs to clearly integrate predictive uncertainty in learning procedure. Consequently, the training examples with large predictive uncertainty will be penalized. We demonstrate the BUP with respect to prediction reliability and out-of-distribution (OOD) predictions. The learned uncertainty is also analyzed in depth. The relations between uncertainty and graph topology, as well as predictive uncertainty in the OOD cases are investigated with extensive experiments. The empirical results with popular benchmark datasets demonstrate the superior performance of the proposed method. 
\end{abstract}

\section{Introduction}
\label{sec:introduction}

\begin{figure}
\centering
\begin{tabular}{c}
\includegraphics[height=3.6cm]{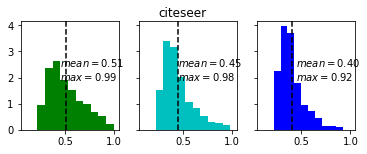}
\end{tabular}
\vspace{-0.2cm}
\caption{Histogram of predictive probability learned with GCN \cite{Kipf2016} for InDist nodes (left), InDist nodes with false predictions (middle), and OOD nodes (right). The x-axis is predictive probability, while the y-axis is its density.}
\label{fig:motivation}
\end{figure}

Modern neural networks have widely been applied in a variety of learning tasks and data modalities due to brilliant performance. However the concern with predictive uncertainty of neural networks has recently been raised \cite{Hasanzadeh2020,Jiang2018,Kendall2017,Guo2017,Depeweg2018,Kumar2019,Elinas2020}, especially in the domains, e.g. healthcare and autonomous driving, where the cost and damage caused by overconfident or underconfident predictions are highly sensitive. 
For graph data, modeling predictive uncertainty is also a critical problem. In node classification, the uncertainty is often represented as predictive probability (confidence), and computed with graph neural networks (GNNs) \cite{Kipf2016,velickovic2018graph,hamilton2017inductive}. Figure~\ref{fig:motivation} visualizes histogram of predictive probabilities, learned with graph convolutional neural network (GCNN) \cite{Kipf2016}, for node classification in the citation network \textit{citeseer}. There are in-distribution (InDist) and out-of-distribution (OOD) nodes. The OOD nodes are nodes from the classes that are not observed in the training data. One can find that the statistics of the inferred predictive probabilities for different types of test nodes are similar. More concretely, the middle panels reveal that the GNN method is confident of its false predictions. The right panels show similar tendency for the OOD case: though the GNN does not learn any knowledge in terms of the OOD nodes (of unobserved classes), it still classifies them to the observed classes with high confidence.
The example demonstrates that in the node classification problem, a more sophisticated manner is needed to properly model the predictive uncertainty.

To meet this challenge, we investigate predictive uncertainty of neural networks for node classification via modeling Bayesian confidence of the predictive probability. In particular, we learn distribution of predictive probability based on message uncertainty. Most existing GNNs pass messages among nodes. Each node is associated with a message, represented as a vector. Through the links, the messages are passed among nodes, such that the information can be shared by the entire network. The messages are often assumed to be deterministic. The GNNs learn a point estimation of each message by minimizing a certain loss, then a predictive probability is computed with e.g.~softmax according to the message. However the messages are \textit{unobservable variables}, introduced as part of the model. Multiple values of the messages could be consistent with the observations. Thus it is reasonable to learn the distributions of the messages, rather than the fixed values of the messages. Technically, we model each message as a random vector following e.g.~a multi-dimensional Gaussian distribution. By integrating uncertainty of messages, we can then learn distributions of predictive probabilities, which allow us to quantify predictive uncertainty in a Bayesian way.  

When the message of a node is transferred to its neighbors, the uncertainty of the message should be passed accordingly. However the commonly used message passing mechanism does not easily apply to uncertainty propagation. Intuitively, predictive uncertainty of a node $i$ should be lower than that of another node $j$ if the node $i$ is connected to more neighbors, since more evidence will flow to the node via the links. However the message passing mechanism behaves reversely, which will lead to larger uncertainty for the well connected nodes, or at least provide no guarantee that the uncertainty will be decreased. To solve the problem, we explore a new uncertainty propagation mechanism based on conditional Gaussian models. The uncertainty of a node $i$ is conditioned on uncertainty of its neighbors. The more neighbors it connects with, the less uncertain its predictive probability. In addition, the more uncertain its neighbors are about the passed messages, the uncertainty of the node will increase accordingly. 
To better integrate predictive uncertainty into the model, we also propose an uncertainty-oriented loss for node classification, that explicitly integrates the predictive uncertainty in the training process. Beyond the commonly used cross entropy loss, the proposed loss penalizes the predictions with high uncertainty. The loss is model agnostic, and can be applied to other classification issues. 
We conduct extensive experiments to test the performance of the proposed method for different cases. We investigate the reliability of the proposed method, and analyze the learned uncertainty in depth.

\section{Related Work}
\label{sec:related}

Uncertainty quantification is essential for practical applications of neural networks \cite{Abdar2021}, which are explored as follows. 

\textbf{Bayesian NNs for Model Uncertainty}
Introducing Bayesian modeling into neural networks (NNs) have attracted a lot of attention in the literature. Instead of learning point estimation of NN parameters, Bayesian methods assume the parameters follow certain prior distributions and learn their posterior to best fit the data. This modeling strategy quantifies the uncertainty of the models themselves (a.k.a.~epistemic uncertainty), and thus often generalize better. Due to computational complexity of Bayesian NNs, a variety of approximation methods have been introduced, e.g., \cite{Kingma2015,Gal2016,Louizos2016}.
In addition, Bayesian NNs are also used to model data uncertainty (a.k.a.~aleatoric uncertainty). For example, \cite{Kendall2017} combined two types of uncertainty with a Bayesian framework for images. \cite{Malinin2018} extended \cite{Kendall2017} to address model uncertainty, data uncertainty and distributional uncertainty. 

\textbf{Non-Bayesian Methods for Uncertainty Estimation}
Bayesian NNs advance the state-of-the-art of uncertainty estimation for deep NNs, however the concern about computational complexity exists. Another line of research is to compute the parameter distributions with non-Bayesian techniques. For example, 
\cite{Osband2016} proposed a bootstrap method and \cite{Lakshminarayanan2017} introduced an ensemble learning method to estimate model uncertainty. 
\cite{Ritter2018} proposed a Kronecker factored Laplace approximation to obtain the posterior of the NN parameters. 

\textbf{Prediction Calibration} This is a notion from frequentist about uncertainty estimation. Unlike Bayesian framework that models randomness of the parameters, the calibration approaches, e.g.~\cite{Chung2021,Minderer2021,Kuleshov2018}, focus on the deviation between the inferred predictions and the empirical long-run frequencies. \cite{Guo2017} presented that the modern deep NNs are not well-calibrated, and analyzed the reasons with extensive experiments. A post-processing calibration method, temperature scaling, was employed to alleviate the miscalibration problem. \cite{Liang2018} extended the work by adding adversary examples to distinguish in- and out-of-distribution images.
\cite{Kumar2019} introduced a scaling-binning calibrator, which combines the ideas of Platt scaling and histogram binning, to reduce sample complexity.
Chung et al.~proposed a quantile based method for calibration \cite{Chung2021}.
\cite{Minderer2021} revisited the calibration issue of recent image classification models, and found that most are among the best calibrated.

\textbf{Bayesian Graph NNs} The aforementioned methods mostly focus on non-graph data, especially images. There are some works that investigated uncertainty in graphs \cite{Eswaran2017,Liu2020,Zhao2020,Monteiro2020,Stadler2021,Chen2022}. In particular, the randomness of the graph structure is explored. For example,
\cite{Zhang2019} combined GCNN and mixed-membership stochastic block model to learn the joint posterior of the random graph (parameters) and the node labels. The method is more robust against noisy links. 
Ma et al.~investigated a flexible generative graph neural network, which models the joint distribution of the node features, labels, and the graph structure \cite{Ma2019}.
\cite{Wang2020} proposed a graph stochastic neural network to learn the distribution of the classification function, and tailored the amortised variational inference to approximate the intractable joint posterior. 
\cite{Hasanzadeh2020} introduced a Bayesian GNN that generalizes stochastic regularization for training.
\cite{Elinas2020} extended the GNNs to the scenarios where no input graph topology is available or there exist noisy edges. 
\cite{Stadler2021} presented a graph posterior network for node classification. 
Chen et al.~\cite{Chen2022} introduced multi-relational graph Gaussian
process network to improve the flexibility of deterministic methods.

\section{Message Passing in GNNs}
\label{sec:message_passing}

There has been a rich literature exploring message passing between nodes with GNNs. A typical example could be the GCNN \cite{Kipf2016}, which defines a kernel to propagate messages:
$
\hat{\mathbf{A}} = \mathbf{A} + \mathbf{I}, \;\;
\mathbf{K} = \mathbf{D}^{-\frac{1}{2}} \hat{\mathbf{A}}
\mathbf{D}^{-\frac{1}{2}} 
$,
where $\mathbf{A}$ denotes adjacency matrix of a graph, and $\mathbf{I}$ is identity matrix corresponding to extra self loops of the nodes. $\mathbf{D}$ is diagonal matrix, which element $d_{i,i}$ is the degree of the node $i$, i.e., $d_{i,i} = \sum\nolimits_j \hat{a}_{i,j}$.
The kernel defines the message passing process:
the message of the node $i$ at the layer $\ell+1$ is the weighted sum of the messages of all directly connected nodes (i.e.~neighbors of $i$, denoted as $N(i)$) at the layer $\ell$. The message $\theta_i^{\ell+1}$ is updated:
\begin{align}
\label{eq:gcn_mess_passing}
\mathbf{\theta}_i^{(\ell+1)} = a( \frac{1}{\sqrt{d_{i,i}}} \sum_{j'\in N(i)} \frac{1}{\sqrt{d_{j',j'}}} \mathbf{\theta}_{j'}^{(\ell)} \mathbf{W}^{(\ell+1)} ).
\end{align}
This message passing mechanism is a deterministic process: no uncertainty/randomness of the messages is considered.

\section{Uncertainty Modeling}
\label{sec:unc_modeling}

Quantifying predictive uncertainty of deep neural networks (DNNs) is an important but yet unsolved problem. The literature has demonstrated that the predictions of DNNs are often overconfident \cite{Nguyen2015,Ovadia2019}. The DNNs may predict wrong labels with high predictive probabilities in many applications.    Fig.~\ref{fig:motivation} illustrates the problem in graph data. Ideally, the predictive probabilities for OOD nodes are low. In fact, the experiments show that the predictive probabilities do not necessarily match it. However the commonly used tools, e.g., confidence interval, do not apply, as labels are discrete variables. \textit{How can we quantify predictive uncertainty for node classification?}

Now we propose a novel method based on Bayesian confidence that will quantify the uncertainty of the prediction via modeling distribution of the predictive probability.  
Typically, GNN classifiers learn class label $\mathbf{y}_i$ of a node $i$ using a discrete probabilistic function with its message $\theta_i$ as input:  
$
p(\mathbf{y}_i|\mathbf{X}, \mathbf{A}) = softmax(\mathbf{\theta}_i), 
\mathbf{\theta}_i = f(\mathbf{X}, \mathbf{A}; \phi),
$
where $\mathbf{y}_i \in \{0,1\}^C $ is a one-hot vector. $\mathbf{X}$ is node features. The function $f(\mathbf{X}, \mathbf{A}; \phi)$ is defined using a GNN (e.g.~GCNN) with $\mathbf{X}$ and $\mathbf{A}$ as inputs, and $\phi$ as parameters. The predictive probability $p(\mathbf{y}_i|\mathbf{X}, \mathbf{A})$ and the message $\theta_i$ are \textit{deterministic} given the function $f$. 
Here we introduce \textit{Bayesian confidence} to acquire extra flexibility. 
In particular, we embed the classifier in a Bayesian framework:
modeling the probabilistic distribution of the predictive probability
$p(\mathbf{y}_i|\mathbf{X}, \mathbf{A})$, and use confidence interval of the probability to derive prediction uncertainty:
\begin{align}
p(\mathbf{y}_i|\mathbf{X}, \mathbf{A}) = \int p(\mathbf{y}_i|\mathbf{\theta}_i) p(\mathbf{\theta}_i|\mathbf{X}, \mathbf{A}) d \theta_i
\end{align}
Here we model the message $\mathbf{\theta}_i \in \mathcal{R}^C$ as a random variable following a distribution $p(\mathbf{\theta}_i|\mathbf{X}, \mathbf{A})$. In Bayesian framework, $\mathbf{\theta}_i$ is known as \textit{unobservable variables}, one for each data example, introduced as part of the models, and will be learned from the data. Multiple values of the message could be consistent with the observations. Thus it is reasonable to model its randomness, i.e.~learning the distribution of the message, rather than a fixed value.
As the message $\mathbf{\theta}_i$ is a continuous variable, we assume it to follow a $C$-dimensional Gaussian distribution:
\begin{align}
p(\mathbf{\theta}_i|\mathbf{X}, \mathbf{A}) &= \mathcal{N}(\mathbf{m}_i,\Sigma_i) \nonumber\\
\mathbf{m}_i = g(\mathbf{X}, \mathbf{A};\gamma)&, \;
\Sigma_i = h(\mathbf{X}, \mathbf{A};\psi)
\label{eq:gaussian}
\end{align}
The covariance matrix $\Sigma_i$ is often assumed to be diagonal. There are mainly two benefits. Firstly, the diagonal covariance simplifies the computation in learning and inference. More importantly, the message $\mathbf{\theta}_i$ is a latent variable, introduced by the model. Each dimension of the latent variable is expected to be independent of each other, and to represent different hidden property of the node. The mean function $\mathbf{m}_i$ and the covariance function $\Sigma_i$ are defined with GNNs, as NNs are good function approximators, especially for complicated data. 
By modeling message uncertainty, the proposed framework can better capture the predictive uncertainty. The larger the variance $\Sigma_i$, the larger the Bayesian confidence interval of the predicted probability. And thus the prediction is more uncertain. Since the message is multi-dimensional, we can use the averaged variance or its Gaussian entropy to quantify the predictive uncertainty. Note that the Gaussian entropy can be viewed as a criterion about the volume of the Gaussian ellipse. We visualize the uncertainty scores in the experiments (see Fig.~\ref{fig:degree_vs_unc}).

In the literature, \cite{Sensoy2018} and \cite{Malinin2018} investigated similar uncertainty, named as \textit{distributional uncertainty}, for image data. They employed Dirichlet instead of Gaussian to model the distribution of the class probability. The Gaussian prior introduced here would be more suitable for graph data, as it explicitly defines uncertainty ($\Sigma_i$) of the message, and can be passed to neighbor nodes with propagation mechanism defined by GNNs.

\section{Uncertainty Propagation}
\label{sec:unc_prop}

We will introduce in the section \textit{how to propagate the uncertainty when passing the messages among the nodes}. A straightforward way could be: learning two GCNNs with the same architecture but different parameters for mean and variance of the Gaussian distribution \eqref{eq:gaussian} respectively, then using the reparameterization trick to draw multiple samples of each message, and employing the MC gradient estimator to approximate the gradients of the parameters for training. However the commonly-used message passing mechanism does not work well when propagating uncertainty. 
Intuitively if a node $i$ is connected with multiple nodes, we can reasonably estimate that prediction uncertainty of the node should be smaller than that of another node $j$ which has less links, since the prediction of the node $i$ will be conditioned on more evidence. The intuition can not be matched with the popular message passing mechanism, e.g.~\eqref{eq:gcn_mess_passing}, where there is no guarantee that linking to more neighbors leads to less uncertainty. 
Thus we introduce a novel propagation mechanism, inspired by Gaussian process (GP) \cite{Rasmussen2006}, to pass uncertainty between nodes. GP defines \textit{a collection of random variables, any finite number of which have a joint Gaussian distribution}. The conditional variance of an example with attributes $\mathbf{x}_*$ is computed as:
$
var(y_*) = k_{*,*} - \mathbf{k}_{*}^T \mathbf{K}^{-1} \mathbf{k}_{*}.
$
$k_{i,j}$ is covariance between the examples $i$ and $j$. The first term of the predictive variance can roughly be understood as the prior variance (uncertainty) of the example, while the second term specifies uncertainty deduction due to dependency on the other examples. 
Inspired by the predictive variance, we introduce the following uncertainty propagation mechanism.

At each layer, we assume the messages of all nodes follow a Gaussian. For a subset of nodes, a node $i$ and its neighbors $N(i)$, each dimension of their messages follow a $(N_i+1)$-dimensional Gaussian, where $N_i$ is the cardinality of $N(i)$. The Gaussian here defines the probabilistic dependencies between nodes, while the Gaussian~\eqref{eq:gaussian} specifies message distribution of a single node. 
Now we designate the covariance.
\begin{definition}
\label{eq:cov}
The covariance matrix between the node $i$ and its neighbors $N(i)$ is a block matrix 
\begin{align}
\left[
\scalemath{0.8}{
\begin{array}{c:cccc}
var(i)  & cov(i,j_1) & cov(i,j_2) & \dots & cov(i,j_{N_i})\\ 
\hdashline[2pt/2pt] 
cov(i,j_1) & var(j_1)  \\
cov(i,j_2) & & var(j_2) & & \text{\large0}\\
\vdots &\text{\large0} & & \ddots \\
cov(i,j_{N_i}) & & & & var(j_{N_i}) 
\end{array}} 
\right]
\end{align}
\begin{align}
cov(i,j) = cor(i,j)\sqrt{var(i)var(j)}, 
cor(i,j) &= 1/\sqrt{\lambda d_{i,i}d_{j,j}} \nonumber
\end{align}
\end{definition}

The definition of the covariance ignores the links between the neighbor nodes, as the links are often sparse in real applications. For dense graphs, it is straightforward to add the covariance $cov(j,j')$ in the matrix without significant influence of the computation of the next steps. In Definition~\ref{eq:cov}, the correlation function $cor(i,j)$ is specified by the subgraph, and thus the covariance matrix can be fully characterized by the variance $var(\cdot)$ of the messages of the node $i$ and its neighbors. The variance is a latent variable that will be learned in training. Here we utilize the correlation function and the learned variance to compute the covariance between the nodes. This will largely decrease the number of the latent variables to be learned. The correlation function can be arbitrary, but must be symmetric. The current one is motivated by the kernel of GCNN, and satisfies the intuition: joint variability of two nodes might be stronger if they have less neighbors, i.e., correlation is inversely proportional to node degrees  $d_{i,i}$. $\lambda$ is a hyperparameter to adjust the correlation. Based on the \textit{Schur complement} Lemma \cite{Zhang2005schur} of block matrix, we can easily prove:
the covariance matrix in Definition~\ref{eq:cov} is valid (the determinant is not zero).
Putting everything together, the conditional variance of the node $i$ is:
\begin{align}
\label{eq:unctrans}
&\hat{var}(i|N(i)) = var(i)-CB^{-1}C^{T}\nonumber\\
&B = diag([var(j_1), var(j_2), \dots, var(j_{N_i})])\nonumber\\
&C = \left[cov(i,j_1), cov(i,j_2), \dots, cov(i,j_{N_i}) \right],
\end{align}
where $B$ and $C$ are blocks in Definition~\ref{eq:cov}. Based on \eqref{eq:unctrans}, we can propagate uncertainty from the neighbors $N(i)$ to the node $i$ at each layer $\ell$.
Across layers, the variations are transferred: 
\begin{align}
var^{(\ell+1)}(i) = a\left( \hat{var}^{(\ell)}(i|N(i)) W_{var}^{(\ell+1)} \right)
\end{align}
With the uncertainty propagation, the more neighbors a node links, the larger the uncertainty is reduced by the available evidence, and thus the smaller the variance of the node is. 

\section{Uncertainty Penalized Loss} 
\label{sec:unc_loss}

Ideally the learned predictive uncertainty should be integrated into the loss to guide the training process. If the prediction of a training example is highly uncertain, then the loss of the example should be penalized. 
However, the commonly-used classification loss functions, e.g.~(expected) cross entropy, do not integrate the  uncertainty properly.
Inspired by the Gaussian likelihood based loss in the regression problem, we introduce a new loss that can directly use message distributions to explicitly integrate the predictive uncertainty into training. 
The new loss is based on Gaussian CDF.
In particular, a training node $i$ is associated with a label $y = c_*$ and a message variable $\mathbf{\theta} = [\theta_{1}, ..., \theta_{C}]$. Each dimension $\theta_{c}$ of the message is independent of each other, and follows a Gaussian distribution with mean $m_{c}$ and variance $\sigma_{c}^2$. As the label of $i$ is $c_*$, we have:
$
\tau_{c} = \theta_{c} - \theta_{c_*}<0, c \neq c_*.
$
$\mathbf{\tau}=[\tau_{1}, ..., \tau_{C}]_{c \neq c_*}^C$ follows a Gaussian with mean $\mu = [m_{1}-m_{c_*}, \ldots, m_{C}-m_{c_*}]$ and covariance:
\begin{align}
\Lambda = 
\begin{bmatrix}
\sigma_{1}^2+\sigma_{c_*}^2 & \sigma_{c_*}^2 &\ldots  &\sigma_{c_*}^2 \\
\sigma_{c_*}^2 & \sigma_{2}^2+\sigma_{c_*}^2 &\ldots  &\sigma_{c_*}^2 \\
 & &\ldots & \\
\sigma_{c_*}^2 & \sigma_{c_*}^2 &\ldots &\sigma_{C}^2+ \sigma_{c_*}^2
\end{bmatrix}
\end{align}
Given the Gaussian, we can now compute the loss $\mathcal{NLL}$ of the example $i$ as Negative Logarithm of its Likelihood probability:
\begin{align}
\label{eq:cdf}
\mathcal{L} = p(y=c_*|\mu,\Lambda) = \int_{-\infty}^0 \mathcal{N}_{C-1}(\mathbf{\tau};\mu,\Lambda) d \mathbf{\tau}
\end{align}
This allows us to compute the likelihood of the discrete label with improper integral of a Gaussian, which avoids the expensive marginalization with unconjugated softmax functions. 
To compute the improper integral, we firstly concert the variable $\tau$ to $\xi$ with Cholesky decomposition $\Lambda = L L^T$ such that the distribution of $\xi$ is a standard Gaussian:
$
\xi = L^{-1}(\tau - \mu) \sim \mathcal{N}_{C-1}(\mathbf{0},\mathbf{I}).
$
Though $\xi_c$ is independent of each other, the improper integral \eqref{eq:cdf} can not be computed in a dimension independent manner, since the integration upper bound of a dimension $c$ is conditioned on the dimensions $c'<c$ before it. 
To solve the issue, we can use a quasi MC approximation based on \cite{Genz1992}, which largely reduces the variance caused by an MC estimator. 
Considering computational efficiency in practice, we can also approximate the likelihood by skipping the off-diagonal entries of $\Lambda$, and the likelihood can be simplified:
\begin{align}
\label{eq:cdf_app}
\mathcal{L} \approx \prod\nolimits_{c=1,c \neq c_*}^C \frac{1}{2} \left[ 1+ erf\left( 
\frac{m_{c_*}-m_{c}}{\sqrt{2(\sigma_{c}^2+\sigma_{c_*}^2)}}
\right)  \right]
\end{align}
where $erf$ denotes error function.
\eqref{eq:cdf_app} clearly incorporates the variance of the message and the difference of the message expectations. The larger the learned mean $m_{c_*}$ at the given label $c_*$ is than the other dimensions $m_c$, the more likely the model makes correct prediction. However, if the model is not sure about the prediction, i.e.~a larger $\sigma_{c}$ or $\sigma_{c_*}$, then the difference $m_{c_*}-m_{c}$ is penalized, the likelihood of the example will change accordingly.

\section{Empirical Analysis}
\begin{table*}
\centering
\caption{Predictive performance of the methods in the normal setting. The metric ACC measures prediction accuracy (the higher the better), while ACE and ECE quantify the reliability of the predictions (the smaller the better). The proposed method BUP is much more reliable than baselines with better or comparable accuracy.}
\label{table:normal}
\vspace{-0.2cm}
\resizebox{\textwidth}{!}{
\begin{tabular}{r|ccc|ccc|ccc|ccc|ccc|ccc} 
\hline
\multirow{2}{*}{Training} & 
\multicolumn{3}{c|}{GCNN}  & \multicolumn{3}{c|}{VGCN} &  
\multicolumn{3}{c|}{BBGDC} & \multicolumn{3}{c|}{BBDE} & 
\multicolumn{3}{c|}{GGP} & \multicolumn{3}{c}{BUP}\\ 
\cline{2-19}
   & ACC & ACE & ECE & ACC & ACE & ECE & ACC & ACE & ECE & ACC & ACE & ECE & ACC & ACE & ECE & ACC & ACE & ECE\\
\hline
Cora 5 & 
\textbf{69.93} & 19.43 & 23.14 &
59.00 & 10.65 & 10.63 &
68.36 & 31.66 & 33.77 &
69.40 & 34.14 & 36.31 &
59.69 & 33.92 & 33.13 &
69.41 & \textbf{9.78} & \textbf{7.82}\\
10 & 
\textbf{75.73} & 19.22 & 21.68 &
64.94 & 12.03 & 11.29 & 
74.52 & 32.85 & 37.02 &
65.75 & 32.66 & 34.82 &
70.47 & 28.66 & 35.32 &
75.02 & \textbf{9.45} & \textbf{6.93}\\
15 & 
\textbf{77.99} & 18.93 & 20.75 & 
68.86 & 11.74 & 10.28 &
74.55 & 32.87 & 37.36 &
76.25 & 35.5 &  40.35 &
68.72 & 23.81 & 29.69 &
77.40 & \textbf{9.90} & \textbf{6.90}\\
20 & 
\textbf{79.10} & 18.80 & 20.06 & 
71.34 & 10.67 & 9.58 &
77.07 & 38.33 & 43.02 &
76.16 & 37.49 & 42.03 &
76.83 & 24.41 & 29.31 &
78.64 & \textbf{10.03} & \textbf{6.80}\\
\hline
Citeseer 5 & 
40.14 & 34.21 & 18.81 &
48.21 & 15.12 & 14.84 & 
50.76 & 14.97 & 13.45 &
44.59 & 19.50 & 19.65 &
41.56 & 29.01 & 20.77 &
\textbf{53.69} & \textbf{10.69} & \textbf{9.93}\\
10 & 
53.75 & 25.63 & 20.58 &
55.62 & 13.22 & 14.20 & 
59.38 & 19.12 & 19.55 &
57.14 & 18.97 & 20.28 &
53.94 & 26.56 & 28.64 &
\textbf{60.38} & \textbf{10.29} & \textbf{8.63}\\
15 & 
63.66 & 18.78 & 20.84 &
58.91 & 13.66 & 14.60 &
\textbf{64.49} & 20.51 & 23.20 &
58.60 & 21.67 & 22.41 &
59.81 & 24.95 & 31.64 &
63.37 & \textbf{10.80} & \textbf{8.36}\\
20 & 
\textbf{67.42} & 17.00 & 20.23 & 
61.03 & 13.49 & 14.94 &
66.80 & 20.62 & 21.09 &
60.12 & 21.17 & 23.48 &
65.27 & 23.09 & 33.07 &
64.77 & \textbf{10.58} & \textbf{8.01}\\
\hline
\end{tabular}}
\end{table*}
Evaluating quality of uncertainty estimation is a challenging task due to the unavailable ground truth of the real uncertainty. To measure the performance of the proposed method BUP, we consider two settings: 
\textit{normal setting} where all test nodes are from the classes observed in the training data and \textit{OOD setting} where the test nodes may belong to an unobserved class. 

\textbf{Datasets} The popular benchmark datasets \textit{cora} and \textit{citeseer} \cite{Elinas2020,Kipf2016,Hasanzadeh2020} are used for the empirical analysis, where the nodes are documents, and features are bag-of-words of the documents. Undirected graphs are constructed according to the citation links. The label of each node specifies the class of the document. The learning task is to predict unknown labels.

\textbf{Baselines} We compare the proposed method with the recent works, including: variational graph convolutional networks (VGCN) \cite{Elinas2020}, Beta-Bernoulli Graph DropConnect (BBGDC and BBDE) \cite{Hasanzadeh2020}, Bayesian semi-supervised learning with graph Gaussian processes (GGP) \cite{ng2018gaussian} and graph convolutional neural network (GCNN) \cite{Kipf2016}. We use  optimal parameter settings and network architectures provided by the authors of VGCN (\url{https://github.com/ebonilla/VGCN}), GGP (\url{https://github.com/yincheng/GGP}), BBGDC/BBDE (\url{https://github.com/armanihm/GDC}), and GCNN (\url{https://github.com/tkipf/gcn}).

\textbf{Metrics} We consider three measurements to quantify the model performance. 
Besides the commonly used predictive accuracy (ACC), we also measure the reliability of the methods with average calibration error (ACE) and expected calibration error (ECE) \cite{Guo2017,Neumann2018}. 
In classification task, a model is reliable (a.k.a.~well calibrated) if the predictive probability is always the true probability. The smaller the scores ACE and ECE are, the closer the two probabilities are. A classifier is expected to achieve good reliability, not just high accurate rate.

\subsection{Q1: Does the proposed method provide more reliable predictions in the normal setting?} 
Here we assume a normal setting, i.e., all classes $\{1,\ldots,C\}$ are observed in the training data. To obtain a comprehensive evaluation, we set different size of training data. In particular, for each data set, We randomly select 5, 10, 15, 20 nodes per class for training, 200 nodes for validation and 2000 nodes for test. We repeat the experiments 10 times, and report the averaged results over all reruns. Table~\ref{table:normal} summarizes the results. One can find that the proposed method achieves superior performance in terms of reliability (ACE and ECE) with better or comparable accuracy rate (ACC). VGCN, as a Bayesian generative neural network, also reports good reliability, although it is still worse than our method BUP. The prediction accuracy of VGCN is not ideal. Other Bayesian methods, such as BBGDC, BBDE and GGP, appear to be less reliable, although they do integrate some uncertainty into the models. GCNN only passes messages between nodes without uncertainty considered. Its reliability is much worse than our method, although GCNN reports slightly higher accuracy in some cases. 
The experiments reveal that our method provides the most reliable predictions with better or comparable accuracy in the normal setting. Learning message uncertainty is an effective way to capture predictive uncertainty.

\subsection{Q2: Does the prediction results of the proposed method report its uncertainty properly in the OOD setting?} 
In the real applications, the OOD setting often happens. For the graph data, we create the OOD examples in the following way. We remove the nodes of the class $C$ from the training and validation, and train the model with the examples from the rest classes $\{1,\ldots,C-1\}$. We denote a node with a label $y_i\in \{1,\ldots,C-1\}$ as an in-distribution example (InDist), and that with $y_i=C$ as an OOD one. For the InDist examples, the predictive performance shows similar tendency as Table~\ref{table:normal}. We show the details in the repository due to the page limit. 

\begin{figure}[!ht]
\centering
\begin{tabular}{c}
\includegraphics[height=3cm]{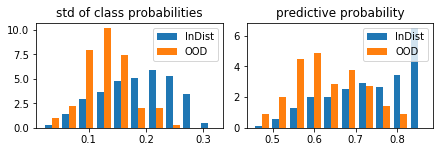}
\end{tabular}
\vspace{-0.5cm}
\caption{Analysis of the inferred predictive probabilities in the OOD setting for the cora data. The distribution of $p_{j,c*}$ for in-distribution nodes (InDist, marked as blue) is significantly different from that of out-of-distribution nodes (OOD, marked as orange). } 
\label{fig:ood}
\end{figure}

Furthermore, we conduct in-depth analysis of the OOD predictions. The experiments are designed as follows. Given a test node $j$, the probability $\mathbf{p}_j$ is $C-1$ dimensional, where $p_{j,c}$ denotes probability of $j$ belonging to $c$. The node $j$ is labeled as $c*$, if $p_{j,c*}>p_{j,c}$, for any $c \in \{1,\ldots,C-1\}$ and $c\neq c*$. The standard deviation (std dev) of $\mathbf{p}_j$ is computed as
$
\sqrt{\frac{1}{C-1}\sum\nolimits_{c=1}^{C-1} \left( p_{j,c} - \frac{1}{C-1}\sum\nolimits_{c=1}^{C-1} p_{j,c} \right)^2}.
$
A smaller std dev means that the predictive probabilities are equally distributed over all dimensions (i.e.~classes), which implies the model is actually not sure which class the test node should belong to. In terms of the maximum probability $p_{j,c*}$, the smaller the probability is, the more dispersed the probability mass. Table~\ref{table:OOD2} reports $p_{j,c*}$ and std dev of $\mathbf{p}_j$ averaged over InDist nodes and and OOD nodes, respectively. One can find that our model reports significantly larger $p_{j,c*}$ for the InDist test nodes, compared with the OOD ones. The std dev of $\mathbf{p}_j$ is much larger for the InDist than for the OOD. These results demonstrate that the proposed method is more confident of the InDist predictions than the OOD ones, and does capture the uncertainty of the OOD nodes. Figure~\ref{fig:ood} illustrates the distributions of $p_{j,c*}$ for the InDist nodes and the OOD ones. The difference is significant, which further visualizes that our method models the uncertainty of the predictions well.
\begin{table}
\centering
\caption{Analysis of the inferred predictive probabilities in the OOD setting for the cora data. Compared with the in-distribution (InDist) nodes, the smaller $p_{j,c*}$ and Std Dev of OOD nodes show the learned probability mass is less concentrated on a single class, but distributed on all possible classes, which implies our method can capture predictive uncertainty correctly.}
\label{table:OOD2}
\vspace{-0.2cm}
\begin{tabular}{c|cc|cc} 
\hline
\multirow{2}{*}{Training} & \multicolumn{2}{c|}{$p_{j,c*}$} & \multicolumn{2}{c}{Std Dev of $\mathbf{p}_{j}$}\\ 
\cline{2-5}
  & InDist & OOD & InDist & OOD\\
\hline
5 &  0.70 &  0.56 &  0.16 & 0.10\\
10 &   0.70 &   0.59 &   0.15 & 0.10\\
15 &   0.75 &   0.62 &   0.19 & 0.13\\
20 &   0.74 &   0.60 &   0.19 & 0.11\\
\hline
\end{tabular}
\end{table}

\subsection{Q3: Can the proposed method capture the influence of the graph structure on the uncertainty of the predictions?}
\begin{figure}
\centering
\begin{tabular}{c}
\includegraphics[height=3.5cm]{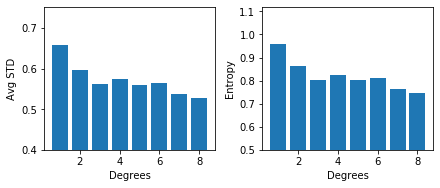}
\end{tabular}
\vspace{-0.3cm}
\caption{Analysis of the learned message uncertainty for the cora data: node degree vs. uncertainty. Our method captures the tendency that the uncertainty (measured with averaged standard deviation and entropy of message distribution) decreases with increasing degrees, as more evidence can be available via neighbors.}
\label{fig:degree_vs_unc}
\end{figure}

Since there is no ground truth about uncertainty, we design this experiment to further validate that the proposed method learns predictive uncertainty appropriately, such that it can capture relations between graph structure and predictive uncertainty.
For example, if a node is connected with multiple neighbors, i.e.~a higher node degree, then its predictive uncertainty is more likely smaller, as more evidence could flow to the node through its neighbors for class prediction. Moreover, if a node is far from the training nodes, then the evidence can vanish during propagation, and thus predictive uncertainty could be higher. We conduct experiments to test whether the learned uncertainty matches the intuitions.
To quantify the uncertainty, we use the learned message distributions. Technically, we select two scores, the averaged standard deviation and the entropy of the message distributions. 
Fig.~\ref{fig:degree_vs_unc} illustrates that the learned uncertainty decreases with the increase of the node degree.
In the repository, we report more results, e.g., the relations between the uncertainty and the averaged shortest path length from the test node to the training nodes. The results reveal that our method BUP does capture the relations between predictive uncertainty and graph structure, and thus is able to learn the predictive uncertainty properly. 

\section{Conclusion}
In this paper, we propose a novel GNN method based on Bayesian confidence of predictive probability to explicitly quantify uncertainty in node classification. Beyond the deterministic process of most GNNs, the proposed method models the uncertainty of the messages, and propagates the uncertainty together with the messages over the entire graph, by which the distribution of the predictive probability can be learned. The extensive empirical analysis demonstrates our method provides much more reliable predictions than state-of-the-art methods.

\bibliographystyle{IEEEtran}
\bibliography{uncertainty}

\begin{thebibliography}{10}
\providecommand{\url}[1]{#1}
\csname url@samestyle\endcsname
\providecommand{\newblock}{\relax}
\providecommand{\bibinfo}[2]{#2}
\providecommand{\BIBentrySTDinterwordspacing}{\spaceskip=0pt\relax}
\providecommand{\BIBentryALTinterwordstretchfactor}{4}
\providecommand{\BIBentryALTinterwordspacing}{\spaceskip=\fontdimen2\font plus
\BIBentryALTinterwordstretchfactor\fontdimen3\font minus
  \fontdimen4\font\relax}
\providecommand{\BIBforeignlanguage}[2]{{%
\expandafter\ifx\csname l@#1\endcsname\relax
\typeout{** WARNING: IEEEtran.bst: No hyphenation pattern has been}%
\typeout{** loaded for the language `#1'. Using the pattern for}%
\typeout{** the default language instead.}%
\else
\language=\csname l@#1\endcsname
\fi
#2}}
\providecommand{\BIBdecl}{\relax}
\BIBdecl

\bibitem{Hasanzadeh2020}
A.~Hasanzadeh \emph{et~al.}, ``Bayesian graph neural networks with adaptive
  connection sampling,'' in \emph{ICML}, 2020.

\bibitem{Jiang2018}
H.~Jiang, B.~Kim, M.~Guan, and M.~Gupta, ``To trust or not to trust a
  classifier,'' in \emph{NeurIPS}, 2018.

\bibitem{Kendall2017}
A.~Kendall and Y.~Gal, ``What uncertainties do we need in bayesian deep
  learning for computer vision?'' in \emph{NIPS}, 2017.

\bibitem{Guo2017}
C.~Guo, G.~Pleiss, Y.~Sun, and K.~Weinberger, ``On calibration of modern neural
  networks,'' in \emph{ICML}, 2017.

\bibitem{Depeweg2018}
S.~Depeweg, J.-M. Hernandez-Lobato, F.~Doshi-Velez, and S.~Udluft,
  ``Decomposition of uncertainty in bayesian deep learning for efficient and
  risk-sensitive learning,'' in \emph{ICML}, 2018.

\bibitem{Kumar2019}
A.~Kumar \emph{et~al.}, ``Verified uncertainty calibration,'' in
  \emph{NeurIPS}, 2019.

\bibitem{Elinas2020}
P.~Elinas \emph{et~al.}, ``Variational inference for graph convolutional
  networks in the absence of graph data and adversarial settings,'' in
  \emph{NeurIPS}, 2020.

\bibitem{Kipf2016}
T.~Kipf and M.~Welling, ``Semi-supervised classification with graph
  convolutional networks,'' in \emph{ICLR}, 2016.

\bibitem{velickovic2018graph}
P.~Velickovic \emph{et~al.}, ``Graph attention networks,'' in \emph{ICLR},
  2018.

\bibitem{hamilton2017inductive}
W.~Hamilton, R.~Ying, and J.~Leskovec, ``Inductive representation learning on
  large graphs,'' in \emph{NIPS}, 2017.

\bibitem{Abdar2021}
M.~Abdar \emph{et~al.}, ``A review of uncertainty quantification in deep
  learning: Techniques, applications and challenges,'' \emph{Information
  Fusion}, 2021.

\bibitem{Kingma2015}
D.~Kingma, T.~Salimans, and M.~Welling, ``Variational dropout and the local
  reparameterization trick,'' in \emph{NIPS}, 2015.

\bibitem{Gal2016}
Y.~Gal and Z.~Ghahramani, ``Dropout as a bayesian approximation: Representing
  model uncertainty in deep learning,'' in \emph{ICML}, 2016.

\bibitem{Louizos2016}
C.~Louizos and M.~Welling, ``Structured and efficient variational deep learning
  with matrix gaussian posteriors,'' in \emph{ICML}, 2016.

\bibitem{Malinin2018}
A.~Malinin and M.~Gales, ``Predictive uncertainty estimation via prior
  networks,'' in \emph{NeurIPS}, 2018.

\bibitem{Osband2016}
I.~Osband, C.~Blundell, A.~Pritzel, and B.~V. Roy, ``Deep exploration via
  bootstrapped dqn,'' in \emph{NIPS}, 2016.

\bibitem{Lakshminarayanan2017}
B.~Lakshminarayanan, A.~Pritzel, and C.~Blundell, ``Simple and scalable
  predictive uncertainty estimation using deep ensembles,'' in \emph{NIPS},
  2017.

\bibitem{Ritter2018}
H.~Ritter, A.~Botev, and D.~Barber, ``A scalable laplace approximation for
  neural networks,'' in \emph{ICLR}, 2018.

\bibitem{Chung2021}
Y.~Chung \emph{et~al.}, ``Beyond pinball loss: Quantile methods for calibrated
  uncertainty quantification,'' in \emph{NeurIPS}, 2021.

\bibitem{Minderer2021}
M.~Minderer \emph{et~al.}, ``Revisiting the calibration of modern neural
  networks,'' in \emph{NeurIPS}, 2021.

\bibitem{Kuleshov2018}
V.~Kuleshov, N.~Fenner, and S.~Ermon, ``Accurate uncertainties for deep
  learning using calibrated regression,'' in \emph{ICML}, 2018.

\bibitem{Liang2018}
S.~Liang, Y.~Li, and R.~Srikant, ``Enhancing the reliability of
  out-of-distribution image detection in neural networks,'' in \emph{ICLR},
  2018.

\bibitem{Eswaran2017}
D.~Eswaran, S.~Günnemann, and C.~Faloutsos, ``The power of certainty: A
  dirichlet-multinomial model for belief propagation,'' in \emph{SDM}, 2017.

\bibitem{Liu2020}
Z.-Y. Liu \emph{et~al.}, ``Uncertainty aware graph gaussian process for
  semi-supervised learning,'' in \emph{AAAI}, 2020.

\bibitem{Zhao2020}
X.~Zhao, F.~Chen, S.~Hu, and J.-H. Cho, ``Uncertainty aware semi-supervised
  learning on graph data,'' in \emph{NeurIPS}, 2020.

\bibitem{Monteiro2020}
M.~Monteiro \emph{et~al.}, ``Stochastic segmentation networks: Modelling
  spatially correlated aleatoric uncertainty,'' in \emph{NeurIPS}, 2020.

\bibitem{Stadler2021}
M.~Stadler \emph{et~al.}, ``Graph posterior network: Bayesian predictive
  uncertainty for node classification,'' in \emph{NeurIPS}, 2021.

\bibitem{Chen2022}
G.~Chen \emph{et~al.}, ``Multi-relational graph representation learning with
  bayesian gaussian process network,'' in \emph{AAAI}, 2022.

\bibitem{Zhang2019}
Y.~Zhang \emph{et~al.}, ``Bayesian graph convolutional neural networks for
  semi-supervised classification,'' in \emph{AAAI}, 2019.

\bibitem{Ma2019}
J.~Ma, W.~Tang, J.~Zhu, and Q.~Mei, ``A flexible generative framework for
  graph-based semi-supervised learning,'' in \emph{NeurIPS}, 2019.

\bibitem{Wang2020}
H.~Wang \emph{et~al.}, ``Graph stochastic neural networks for semi-supervised
  learning,'' in \emph{NeurIPS}, 2020.

\bibitem{Nguyen2015}
A.~Nguyen \emph{et~al.}, ``Deep neural networks are easily fooled: High
  confidence predictions for unrecognizable images,'' in \emph{CVPR}, 2015.

\bibitem{Ovadia2019}
Y.~Ovadia \emph{et~al.}, ``Can you trust your model’s uncertainty? evaluating
  predictive uncertainty under dataset shift,'' in \emph{NeurIPS}, 2019.

\bibitem{Sensoy2018}
M.~Sensoy, L.~Kaplan, and M.~Kandemir, ``Evidential deep learning to quantify
  classification uncertainty,'' in \emph{NeurIPS}, 2018.

\bibitem{Rasmussen2006}
C.~Rasmussen and C.~Williams, \emph{Gaussian Processes for Machine
  Learning}.\hskip 1em plus 0.5em minus 0.4em\relax MIT Press, 2006.

\bibitem{Zhang2005schur}
F.~Zhang, Ed., \emph{Schur Complement and Its Applications}.\hskip 1em plus
  0.5em minus 0.4em\relax Springer, 2005.

\bibitem{Genz1992}
A.~Genz, ``Numerical computation of multivariate normal probabilities,''
  \emph{Journal of Computational and Graphical Statistics}, 1992.

\bibitem{ng2018gaussian}
Y.~C. Ng, N.~Colombo, and R.~Silva, ``Bayesian semi-supervised learning with
  graph gaussian processes,'' in \emph{NeurIPS}, 2018.

\bibitem{Neumann2018}
L.~Neumann \emph{et~al.}, ``Relaxed softmax: Efficient confidence
  auto-calibration for safe pedestrian detection,'' in \emph{NIPS Workshop},
  2018.

\end{thebibliography}

\end{document}